\documentclass[letterpaper, 10 pt, conference]{ieeeconf}  

\IEEEoverridecommandlockouts                              

\usepackage{graphics} 
\usepackage{graphicx}
\usepackage{epsfig} 
\usepackage{times} 
\usepackage{amsmath} 
\usepackage{amssymb}  
\usepackage[font=small]{caption}
\usepackage{subcaption}

\usepackage{fontenc}
\usepackage{mathtools}
\usepackage{color}
\usepackage{citesort}
\usepackage[bottom]{footmisc}
\usepackage{epstopdf}
\usepackage{tabularx}
\usepackage{hhline}
 \usepackage{algorithm}
\PassOptionsToPackage{noend}{algpseudocode}
\usepackage{algpseudocode}

\definecolor{grey_color}{RGB}{84,84,84}

\DeclareMathAlphabet{\pazocal}{OMS}{zplm}{m}{n}

\newcommand{\Ns}{\pazocal{N}}

\newcommand{\As}{\pazocal{A}}

\newcommand{\Cs}{\pazocal{C}}
\newcommand{\Ds}{\pazocal{D}}
\newcommand{\Gs}{\pazocal{G}}

\newcommand{\Rs}{\pazocal{R}}
\newcommand{\Es}{\pazocal{E}}

\newcommand{\Vs}{\pazocal{V}}

\newcommand{\Ss}{\pazocal{S}}

\newcommand{\Ps}{\pazocal{P}}
\newcommand{\ps}{\mathbf{p}}
\newcommand{\ds}{\mathbf{d}}

\newcommand{\threeD}{$3\textrm{D}$}

\newcommand{\norm}[1]{\left\lVert#1\right\rVert}

\title{\LARGE \bf
Distributed Infrastructure Inspection Path Planning subject to Time Constraints
}

\author{Kostas Alexis, Christos Papachristos, Roland Siegwart, Anthony Tzes
}

\begin{document}

\maketitle
\thispagestyle{empty}
\pagestyle{empty}

\begin{abstract}
Within this paper, the problem of \threeD~structural inspection path planning for distributed infrastructure using aerial robots that are subject to time constraints is addressed. The proposed algorithm handles varying spatial properties of the infrastructure facilities, accounts for their different importance and exploration function and computes an overall inspection path of high inspection reward while respecting the robot endurance or mission time constraints as well as the vehicle dynamics and sensor limitations. To achieve its goal, it employs an iterative, $3$--step optimization strategy at each iteration of which it first randomly samples a set of possible structures to visit, subsequently solves the derived traveling salesman problem and computes the travel costs, while finally it samples and assigns inspection times to each structure and evaluates the total inspection reward. For the derivation of the inspection paths per each independent facility, it interfaces a path planner dedicated to the \threeD~coverage of single structures. The resulting algorithm properties, computational performance and path quality are evaluated using simulation studies as well as experimental test--cases employing a multirotor micro aerial vehicle.
\end{abstract}

\section{INTRODUCTION}\label{sec:intro}

Aerial robotics are getting integrated into a wide variety of critical applications. Among the most promising ones, is that of infrastructure inspection and maintenance operations~\cite{Metni20073,englot2012sampling,ADBS_AURO_2015,InfrInspHeli,LocLinearStruInsp,DABS_ICRA_14,BABOOMS_ICRA_15,NBVP_ICRA_16,Oettershagen_FSR2015,ZPAT_ISVC_2015,SIP_AURO_2015,APST_MSC_2015,bircher_robotica}. Aerial robots correspond to a safety--ensuring, efficiency--improving and cost--saving asset that can revolutionize this field. Within such missions, one of the most challenging problems -- alongside those of perception, estimation and control -- is that of being able to autonomously derive the inspection path which handles the structural and spatial distribution properties of the infrastructure, is efficient and respects the often tight endurance limitations of the vehicle as well as any sensor-- or kinematic--constraints that apply. Especially in the case of distributed infrastructure --such as wind farms, solar panels, oil rigs or the power network-- the complexity of this problem is particularly high. 

The work presented in this article augments in terms of results the paper in~\cite{papachristos2016distributed} and deals exactly with the problem of distributed infrastructure inspection path planning for aerial robotics that are subject to time constraints. The new algorithm considers the fact that given a set of spatially distributed \textit{Infrastructure Facilities of Interest} (IFIs) and a robot with time constraints, intelligent inspection path planning should aim to derive the best \textit{possible} path and not be constrained on a potentially infeasible attempt to find a full--coverage solution. Therefore, the proposed algorithm considers an \textit{exploration function} (EF) for each IFI as well as an \textit{importance weight} (IW) and aims to compute the path that maximizes the totally collected \textit{Inspection Rewards} (IRs) (combination of IWs and EFs) while respecting the vehicle endurance or other mission time constraints, as well as the vehicle motion and sensor limitations. It is highlighted that the algorithm is free to select the partial structural inspection of a subset of the spatially distributed IFIs as long as this maximizes the totally collected IR given that the imposed time constraints do not allow complete inspection of all IFIs. To the authors best knowledge, there is only a small body of research activities trying to address this problem in an autonomous manner despite its importance for robotic inspection operations. 

%
\begin{figure}[h!]
\centering
  \includegraphics[width=0.99\columnwidth]{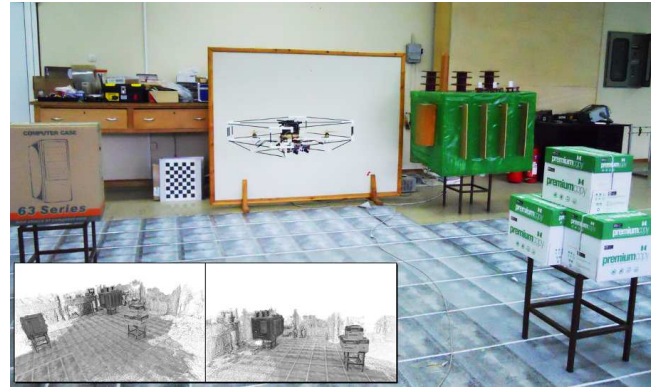}
\caption{Photo of an instant of a distributed structural facilities inspection mission using an autonomous aerial robot. Views of the reconstructed point cloud are also depicted. }
\label{fig:intro_photo}
\end{figure}
%

Subject to specific limiting assumptions, this problem shares certain similarities with the Orienteering Problem (OP) class. Considering constant, time independent, inspection rewards it can be solved using methods for OP~\cite{vansteenwegen2011orienteering,tsiligirides1984heuristic,blum2007approximation}. Assuming a time--only dependent and nondecreasing learning curve, the problem shares similarities with the Traveling Tourist Problem (TTP) and the planning of optimal tourist itineraries given trip time constraints~\cite{stanley2001statistical,gavalas2015heuristics,yu2014optimal,con_dom_sets}. However, in practice the exploration function for a structure is not a single function of time since two equal duration inspection paths that start at different points or follow different directions can lead to largely different coverage percentages. This fact further increases the complexity of the problem. The proposed algorithm is specifically designed to be able to deal with this situation and employs an iterative $3$--step optimization paradigm, within which it randomy samples a subset of the facilities that should be visited and partially or fully inspected \textit{(first step)}, subsequently solves the derived Traveling Salesman Problem (TSP) and computes the travel costs \textit{(second step)}, and finally \textit{(third step)} samples and assigns inspection times and subsets of the full--coverage path of each IFI that respect the time constraints and then evaluates the overall gained inspection reward. The computation of the full--coverage path per IFI is achieved by interfacing another inspection path planner that is capable of computing optimized inspection paths for a single structure. Any relevant path planner can be employed for this task, while within the framework of this work, an algorithm previously proposed and open--sourced by one of the authors is employed~\cite{BABOOMS_ICRA_15}. 

The proposed approach is thoroughly analyzed using both simulation as well as experimental studies that employ the autonomous aerial robot shown in Figure~\ref{fig:intro_photo}. Computational analysis is provided with a detailed description of the breakdown of the computational cost for each main algorithmical step. The collected dataset from the experimental studies and a large family of simulation results are publicly released in order to allow easier future comparison with other methods and strategies proposed by the research community~\cite{DISIPdatasest}. This algorithm will also be open--sourced including the interfaces to the employed single structure inspection planner with the aim to release a public and complete infrastructure inspection path--planning framework.

The distributed structural inspection path planning problem is defined in Section~\ref{sec:problem}, followed by the detailed description of the proposed approach in Section~\ref{sec:approach}. Finally, evaluation studies are presented in Section~\ref{sec:evaluation}, while conclusions are drawn in Section~\ref{sec:concl}.

\section{PROBLEM DEFINITION}\label{sec:problem}

The problem of distributed infrastructure structural inspection path planning subject to time constraints, as considered in this paper, consists of a) a set of spatially distributed \textit{Infrastructure Facilities of Interest} (IFIs) that are modeled using $3\textrm{D}$ meshes, and each of them is associated with an \textit{Inspection Reward} (IR) that is computed as the multiplication of the \textit{Importance Weight} (IW) and the \textit{Exploration Function} (EF), b) a time constraint (either due to the endurance of the robot or mission--specific limits), c) dynamic constraints of the vehicle as well as d) sensor Field--of--View (FoV) limitations. Assuming that for each IFI, an \textit{admissible} full--coverage path exists and can be computed using a dedicated Single Structure Inspection Path--planner (SSIP), the goal is to maximize the total collected IRs by finding the best combination of a subset of IFIs to be visited, associated inspection times for each IFI and corresponding subsets of its full--coverage inspection path, as well as the tour among the selected IFIs. Note again, that the problem of SSIP is decoupled and addressed by interfacing a relevant solver such as the one previously proposed by one of the authors~\cite{BABOOMS_ICRA_15}. In the following, the addressed problem is defined more formally while the basic notation is also introduced.

Let $\Gs = \{\Vs,\As\}$ be a graph, where $\Vs = \{v_1,v_2,...,v_{N}\}$ is the set of vertices each one corresponding to the couple of entry and exit pose configurations of the coverage path of one IFI $3\textrm{D}$ structure $\Ss_i \in \Ss$ and $\As$ is the corresponding arc set. Note that incoming connections to a vertex $v_i$ are attached to the entry point of the coverage path of the IFI $\Ss_i$ while outbound connections start from the exit point of the coverage path. Furthermore, $\Rs = \{r_i(\ps_i^{\alpha,T_i,\Ss_i})\},\forall~\Ss_i \in \Ss,~r_i(\ps_i^{\alpha,T_i,\Ss_i}) = w_i f_i(T_i,p_i^\alpha)$ is the set of IR functions for each of the IFIs vertices with $w_i$ being the IW and $f_i(T_i,p_i^\alpha)$ is the EF as a function of the inspection time $T_i$ and starting point $p_i^\alpha$ in the full--coverage path $\ps_i(\Ss_i)$ of $\Ss_i$, and $t_{ij}$ is the transition time associated with each arc $a_{ij} \in \As$.  The subset of the full--coverage path $\ps_i(\Ss_i)$ that starts from $p_i^\alpha$ and has a duration $T_i$ for a maximum travel speed $\upsilon_T$ is denoted as $\ps_i^{\alpha,T_i,\Ss_i}$. Now let a feasible solution to the problem be the 5--tuple  $\Ds = \{\Vs^\prime,\As^\prime,\Ss^\prime,\Ps^\prime,\Rs^\prime\}$, where $\Gs^\prime = \{\Vs^\prime,\As^\prime\}$ denotes the hamiltonian path among the sampled vertices $\Vs^\prime \subseteq \Vs$ associated with the sampled IFIs set $\Ss^\prime \subseteq \Ss$ and the arc set $\As^\prime$, $\Ps^\prime =\{\ps_i^{\alpha,T_i,\Ss_i}\},~\forall \Ss_i \in \Ss^\prime$ are the sampled subsets of the full--coverage paths for the selected IFIs and $\Rs^\prime = \{r_i(\ps_i^{\alpha,T_i,\Ss_i})\},~\forall \Ss_i \in \Ss^\prime$ is the relevant set of inspection rewards. The optimal solution to the problem is the 5--tuple $\Ds_{OPT} = \{\Vs^\prime_{OPT},\As^\prime_{OPT},\Ss^\prime_{OPT},\Ps^\prime_{OPT},\Rs^\prime_{OPT}\}$ such that:

\small
\begin{eqnarray}
 \mathbf{\max} R_{TOT},~~R_{TOT} = \sum_{i: \Ss_i \in \Ss^\prime} r_i(\ps_i^{\alpha,T_i,\Ss_i}) \\
\textrm{s.t.}~ T_{TOT} \le T_{\max},~~T_{TOT} = \sum_{i: \Ss_i \in \Ss^\prime}(T_i) + \sum_{i,j: \Ss_i,\Ss_j \in \Ss^\prime} t_{ij}
\end{eqnarray}
\normalsize
Note that not only the aforementioned time constraint of the mission should be respected but also the dynamic constraints of the vehicle and the limitations of the sensor as part of the selected SSIP calculations and the Boundary Value Solver (BVS) that is employed for the vehicle dynamics and will be explained in the next section. 

\section{PROPOSED APPROACH}\label{sec:approach}

The proposed Distributed Infrastructure Structural Inspection Planner \textit{(DISIP)} is overviewed within this section. It relies on an iterative, $3$--step optimization strategy that allows it to compute inspection paths that improve the gained inspection reward while respecting the imposed time constraints. Within each iteration, the three steps executed are: a) random sampling of a subset of the possible IFIs to be inspected, b) solution of the derived TSP problem and finally c) randomized assignment of inspection times (up to the limit of the time constraint) and corresponding subsets of the full--coverage path for each sampled IFI. 
As this procedure runs iteratively, the algorithm manages to provide improved solutions over the course of time, while a first solution is available from the first run. These steps are summarized in Algorithm~\ref{alg:DISIP_algorithm} while the most important functions are detailed in the subsequent sections.

\begin{algorithm}[h]
\caption{Distributed Infrastructure Structural Inspection Planner (DISIP)}
\label{alg:DISIP_algorithm}
\begin{algorithmic}[1]
\State $k \leftarrow 0$
\State Computation of SSIP paths for all IFIs (Section \ref{sssec:SSIP})
\State Inspection Structures Set Sampling (Section \ref{sssec:ISS})
\State Cost matrix computation (Section \ref{sssec:costmat})
\State Solve the TSP problem using the LKH to obtain initial tour (Section \ref{sssec:TSP})
\State Inspection Times SSIP paths Sampling (Section \ref{sssec:ITS})
\State Inspection Rewards computation (Section \ref{sssec:award})
\While{running}
	\State{Resample Inspection Structures Set (Section \ref{sssec:ISS}) }
	\State{Recompute the cost matrix (Section \ref{sssec:costmat})}
	\State{Recompute the best tour among the sampled structures (Section \ref{sssec:TSP})}
	\State{Resample Inspection Times and SSIP paths (Section \ref{sssec:ITS})}
	\State{Recompute Inspection Rewards and update best tour reward $R_{best}$ and best solution $\Ds_{best} = \{\Vs_k^\prime,\As_k^\prime,\Ss_k^\prime,\Ps_k^\prime,\Rs_k^\prime\}$ is applicable (Section \ref{sssec:award})}
	\State{$k\leftarrow k+1$}
\EndWhile
\State{\textbf{end while}} 
\State{Assemble the best DISIP path $\ds_{best}$ (Section \ref{sssec:disip_assem})} \\
\Return{$\Ds_{best}$, $\ds_{best}$, $R_{best}$}
\end{algorithmic}
\end{algorithm}

\subsection{Computation of the SSIP paths}\label{sssec:SSIP}

The DISIP algorithm utilizes and interfaces a Single Structural Inspection Path--planner (SSIP) that is capable of computing a complete coverage path for a given connected structure as well as a sensor model and vehicle kinematic constraints. Given the maximum travel speed for inspection $v_T^I$ and max yaw rate $\dot{\psi}_{\max}$, the computed full coverage path is timed at each step. Different SSIP algorithms can be used, while within this work the algorithm previously proposed by one of the authors~\cite{BABOOMS_ICRA_15} is employed both due to its performance as well as due to its open--source availability. The algorithm is executed once per IFI type, considers a triangular mesh representation of the structure and computes the path via an optimization method that alternates between two steps, namely a step that samples and computes new viewpoints in order to reduce the cost--to--travel between each viewpoint and its neighbours, and a step that computes the optimal connecting tour for the current iteration. The specific planner supports both rotorcraft as well as fixed--wing aerial vehicles and accounts for their different motion model. Furthermore, its computational cost is particularly small due to the convexification of the viewpoint computation problem and the utilization of the Lin--Kernighan--Helsgaun (LKH) solver for the tour computation step. Computational analysis is included in the relevant publication~\cite{BABOOMS_ICRA_15}, while demo scenarios and experimental datasets are available online. 

\subsection{Inspection Structures Set Sampling}\label{sssec:ISS}

Within the IFIs set $\Ss$ and the associated graph vertices $\Vs$, the subsets $\Ss_k^\prime \subseteq \Ss$, $\Vs_k^\prime \subseteq \Vs$ are randomly sampled at each iteration $k$. The rest of the steps of the DISIP algorithm are now executed based on this set of IFIs and the corresponding graph $\Gs_k = \{\Vs_k^\prime,\As_k\}$, where $\As_k$ is the set of all possible connections between the members of $\Vs_k^\prime$. 

\subsection{Cost Matrix Computation}\label{sssec:costmat}

The paths among all $v_i \in \Vs_k^\prime$ are computed via the utilization of a two state BVS applicable to rotorcraft aerial robots. This BVS is employed to connect two $v_i,v_j$ based on the start and end points of the corresponding sampled inspection path of the two IFIs $\Ss_i,\Ss_j$ respectively (asymmetric connections). It is noted that the employed BVS of a rotorcraft aerial robot consists of position as well as yaw $\xi = \{x,y,z,\psi\}$, while as long as low speeds are considered, roll and pitch are approximated to be zero. Consequently, the path from configuration $\xi_i$ to $\xi_j$ is given by $\xi(s) = s\xi_1 + (1-s)\xi_0$, where $s\in[0,1]$. The two limitations considered are the maximum translational velocity constraint $\upsilon_{T}$ and the yaw rate constraint $\dot\psi_{\max}$. The resulting execution time is $t_{ij} = \max(\delta/\upsilon_{T},\norm{\psi_j - \psi_i}/\dot{\psi}_{\max})$ where $\delta$ denotes the Euclidean distance. The cost of a path segment within the DISIP algorithm execution corresponds to its travel--execution time $t_{ex}$. This calculation is conducted for all possible connections between the sampled IFIs of any iteration and the corresponding cost matrix $\Cs_k$ is derived. As a remark, it is noted that a local collision--free point--to--point navigation algorithm such as RRT$^\star$ may be used to avoid collision with the IFI from which a connection departs, arrives or other IFIs in between.

\subsection{TSP Tour Solution}\label{sssec:TSP}

Given the cost matrix $\Cs_k$, the next goal is to compute the hamiltonian path $\Gs_k^\prime = \{\Vs_k^\prime,\As_k^\prime \}$ where $\As_k^\prime$ is the arc set of the hamiltonian path. Essentially, at this step the algorithm derives the solution to the (possibly asymmetric) TSP problem of visiting the sampled vertices $\Vs_k^\prime$ of the sampled IFIs set $\Ss_k^\prime$. Among the multiple algorithms that have been proposed for the TSP, the approach of Lin--Kernighan~\cite{lin1973effective} and its methodology of implementation by Keld Helsgaun (LKH solver)~\cite{helsgaun2000effective} corresponds to the best known local search algorithm solution. The LKH solver relies on the concept of $\lambda$--optimality, according to which a tour is said to be $\lambda$--optimal (or simply $\lambda$--opt) if it is impossible to obtain a shorter tour by replacing any $\lambda$ of its links by any other set of $\lambda$ links. Similarly, a $\lambda$--opt neighborhood for tour $\chi$, $\Ns_k ^{\lambda}(\chi)$, consists of all tours which can be constructed by deleting and adding $\lambda$ edges. Based on the observation that two hamiltonian cycles only differ in $\lambda$ edges ($2 \le \lambda \le N_k$) ($N_k$ the sampled number of IFIs at the $k$--th iteration), i.e. $\chi \in \Ns _{k}^{\lambda} (\chi),~\forall \chi$ and in order to address the problem that $\lambda$--optimality can be tested in $\pazocal{O}(N_k^\lambda)$ while $\lambda$ is unknown, LKH employs the alternative of choosing an \textit{efficient searchable} neighborhood such that $\lambda$ can be chosen dynamically. This is the concept of \textit{sequential} $\lambda$--opt \textit{moves}. More formally, a $\lambda$--opt move is called sequential if it can be described by a path alternating process between deleted and added edges. The code implementation of the LKH solver is found online at~\cite{LKHcode}. Using such a solver, the derived TSP problem from the sampled graph $\Gs_k^\prime$ is computed, leading to the tour $\tau_k$ (the solution to the problem of finding the hamiltonian path $\Gs_k^\prime =\{\Vs_k^\prime,\As_k^\prime\}$) which represents the sequence based on which the sampled IFIs should be visited as well as the travel costs of this route $T^{\tau_k}$.

\subsection{Inspection Times and SSIP paths Sampling}\label{sssec:ITS}

For each IFI $\Ss_i$, a SSIP path $\ps_i(\Ss_i)$ is computed, has $m_i$ number of points and $m_i-1$ path segments, traveled in time based on the same BVS described above. Given the overall SSIP path duration $T_i$ of each IFI, an inspection time $T_i^k \le T_i$ for each IFI $\Ss_i \in \Ss_k^\prime$ and DISIP iteration $k$ is sampled while accounting that the overall assigned time $\sum_{i:~\Ss_i \in \Ss^\prime} T_i^k$ should be less than the available time $T_{\max}-T^{\tau_k}$. As by random sampling, a set of $T_i^k$ that would overall lead to $\sum_i T_i^k = T_{\max} - T^{\tau_k}$ exactly is unlikely, an additional step of adjusting the assigned times by an amount proportional to the ratio of the sampled $T_i^k$ and leads to the satisfaction of the aforementioned equation takes place. Based on the $T_i^k$ values, the algorithm further samples an entry $p_i^{\alpha_k}$ point to each of the SSIP paths $\ps_i(\Ss_i)$ as well as an exit point $p_i^{\beta_k}$ such that the corresponding subset of the SSIP path $\ps_i^{\alpha_k,T_i^k,\Ss_i}$ has an overall duration time as close as possible (given the discrete points of the SSIP path) to $T_i^k$. Minor adjustments (interpolation) may take place afterwards to account for the discretized nature of the SSIP paths.

\subsection{Inspection Reward Computation}\label{sssec:award}

For a SSIP path $\ps_i(\Ss_i)$, each point corresponds to a viewpoint configuration that covers a specific subset of the IFI structure model (one of its triangular mesh faces). Therefore, for the sampled subset of each SSIP path $\ps_i^{\alpha_k,T_i^k,\Ss_i}$, the overall covered area of the IFI $\Es_i^k$ can be computed,	 which then allows the computation of the coverage ratio $\gamma_i^k = \Es_i^k /\Es_i$ (where $\Es_i$ the overall area of the $i$--th IFI). Once this value is computed for all sampled inspection paths for the given set of IFIs within the $k$--th iteration, the inspection reward of each path is derived as $r_i^k(\ps_i^{\alpha_k,T_i^k,\Ss_i}) = w_i \Es_i^k$. Finally, the total inspection reward is computed as $R_{TOT}^k = \sum_{i: \Ss_i \in \Ss^\prime_k} r_i^k(\ps_i^{\alpha_k,T_i^k,\Ss_i})$. If this reward is higher than the previous best value $R_{best}$, then the $k$--th iteration of the algorithm is considered as the so--far best solution and the reward value gets updated $R_{best} \leftarrow R_{TOT}^k$. Furthermore, the elements of this best DISIP solution $\Ds_{best} = \{\Vs_k^\prime,\As_k\prime,\Ss_k\prime, \Ps_k\prime,\Rs_k\prime\}$ are exported. Note that this 5--tuple contains the information for the best TSP tour $\tau_{best} \leftarrow \tau_k$ within the hamiltonian path $\Gs_k^\prime = \{\Vs_k^\prime,\As_k^\prime\}$.

\subsection{Assembly of the DISIP path}\label{sssec:disip_assem} 

At this final step, which is performed after the execution of all the iterations of the DISIP algorithm, the SSIP paths $\Ps_k^\prime$ and the connecting segments of the TSP tour $\tau_{best}$ are combined to form the overall DISIP path. This connected timed path $\ds_{best}$ is the result used for the guidance of the aerial robot. 

\subsection{DISIP Performance}

The proposed solution to the problem of distributed infrastructure structural inspection path planning subject to time constraints is motivated from the need to provide high--quality inspection paths that respect all the constraints, are computed fast, while given that more time is available for the mission preparation, updated and improved solutions can be found. The algorithm does not possess the feature of optimality, it however produces high quality paths that make use of all the time that is available to the robot even from the first solution. As each iteration requires a very short computation time, its iterative execution quickly leads to paths of very high quality. Furthermore, the algorithm makes no constraining assumption on the nature of the exploration function and uses any possible full--coverage path for each of the distributed structures from which it extracts subsets based on the sampled time assignments. To the authors best knowledge this is the first attempt of its kind while the provided simulation and experimental studies further demonstrate its efficiency and suitability for the problem.

\section{EVALUATION STUDIES}\label{sec:evaluation}

In order to enable thorough evaluation of the properties and performance of the proposed distributed infrastructure inspection path planner, a wide set of evaluation test--cases both in simulation as well as using experimental studies were considered. Within those, the goal is to assess the quality of the inspection path for different spatial sparsity and distribution of the infrastructure as well as different robot endurances or mission time constraints and finally the computational properties of the algorithm. The experimental studies were conducted using a quadrotor MAV that is capable of autonomous navigation using a Visual--Inertial localization and mapping framework~\cite{PTT_IROS_15}. 

\subsection{Studies in Simulation}\label{subsec:sim}

In order to evaluate the properties of the DISIP algorithm and especially the quality of the inspection paths for different levels of spatial sparsity of the IFIs, we followed a process of creating multiple randomized distributed infrastructure scenarios. All scenarios consist of a combination of four ($4$) different infrastructure facilities and more specifically, a) a solar--panels park, b) a vessel tank, c) a turbine storage and a d) distribution network power transformer. These structures correspond to a variety of sizes and inspection complexities. Having these four building blocks at hand, $40$ simulations were considered and specifically $10$ for each case of $8, 16, 32, 64$ of these structures (randomly selected) at an overall area with planar dimensions $[\Delta x, \Delta y, \Delta z] = [200, 200, 50]\textrm{m}$ (for $8$ or $16$ structures) or $[\Delta x, \Delta y, \Delta z] = [400, 400, 100]\textrm{m}$ (for $32$ or $64$ structures). As the same size of overall area was used for different amount of structures, this also allowed to evaluate different infrastructure densities. Furthermore, it is pointed out that for the smaller scenarios, an aerial robot with an endurance of $T_{\max} = 1800\textrm{s}$, a traveling speed $v_T = 3\textrm{m/s}$, and inspection speed $v_T^I = 1\textrm{m/s}$ and a yaw rate constraint $\dot{\psi}^{\max} = 0.5\textrm{rad/s}$ was considered (values achievable by a large multirotor), while for the larger scenarios the endurance was set to $T_{\max} = 5400\textrm{s}$ (achievable by a gasoline--powered helicopter UAV). Regarding, the assumed on--board camera sensor, all scenarios were conducted assuming a FOV of $65^\circ$ in both dimensions. Without loss of generality, all importance weight (IW) factors except those refering to the solar panel park are set to one, while for the solar panel farm it is set to $0.25$.

\begin{figure}[h!]
\centering
  \includegraphics[width=0.99\columnwidth]{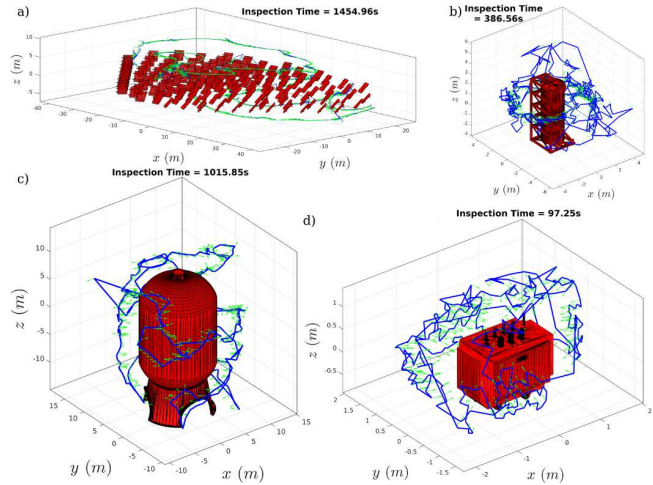}
\caption{Results of the execution of the SSIP algorithm for each structure. The DISIP algorithm interfaces SSIP and uses these full--coverage path to extract a subset of them based on the time assigned for each structure and the expected inspection reward.  }
\label{fig:SSIPresults}
\end{figure}

The first step of the DISIP algorithm for all these scenarios is to interface SSIP and compute the full--coverage inspection paths for each structure present at the scenario. Figure~\ref{fig:SSIPresults} presents these coverage paths for all the considered structures possibly available in the aforementioned scenarios. 

%
%

Having the SSIP results at hand, DISIP can proceed to its further calculations as described in Section~\ref{sec:approach}. In the following, three simulation scenarios will be presented in detail, while summarized statistical data from all the $40$ simulations are provided subsequently. The complete set of simulation results data may be found online~\cite{DISIPdatasest}.

\begin{figure}[h!]
\centering
  \includegraphics[width=0.99\columnwidth]{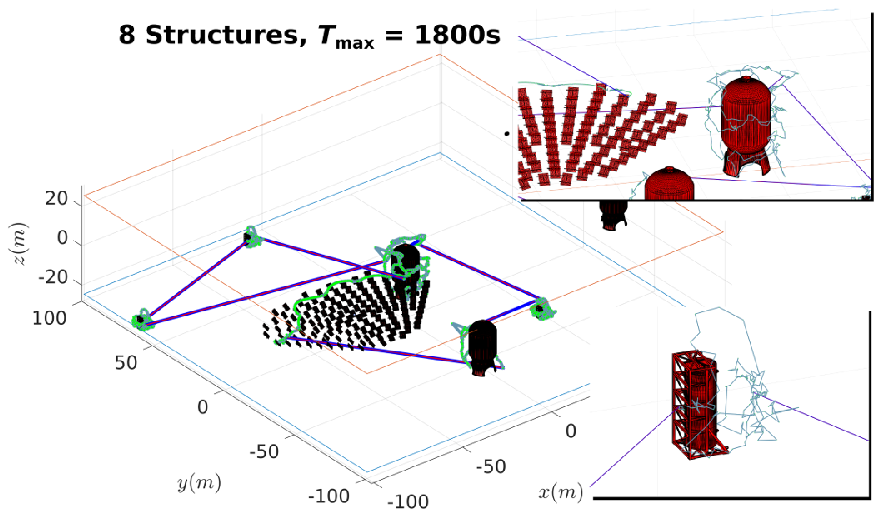}
\caption{Distributed infrastructure inspection result using the proposed algorithm for the case of a large area with $8$ structures and a robot endurance of $T_{\max} = 1800\textrm{s}$. The distribution of the structures is based on a randomized spatial generator.}
\label{fig:DISIP_8}
\end{figure}

\begin{figure}[h!]
\centering
  \includegraphics[width=0.99\columnwidth]{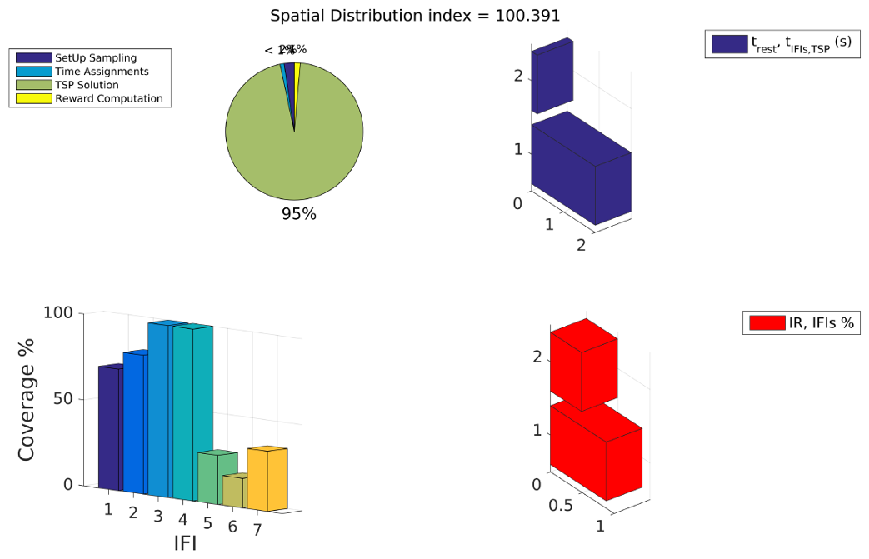}
\caption{Computational analysis of the a distributed infrastructure inspection result using the proposed algorithm for the case of a large area with $8$ structures and a robot endurance of $T_{\max} = 1800\textrm{s}$. The term spatial distribution index denotes the average distance from IFI to IFI. The upper right plot denotes the time for the IFI sampling and TSP computations (lower bar) and all the rest (upper bar). The bottom right plot denotes the percentage of IFIs visited (lower bar) and the percentage of awards collected (upper bar). The upper left pie denotes the percentage of the basic algorithmical step and as shown the TSP solution consumes the majority of the time. Finally, the bottom left plot denotes the percentage of coverage for each of the sampled IFIs. }
\label{fig:DISIP_8_Analysis}
\end{figure}

The first result to be analytically presented, refers to the case of an aerial robot with $T_{\max} = 1800\textrm{s}$ endurance and its other parameters as previously mentioned, while a scenario consisting of $8$ IFIs is considered. The best computed inspection result, after $k=30$ DISIP iterations,  is shown in Figure~\ref{fig:DISIP_8}, while statistical analysis is shown in Figure~\ref{fig:DISIP_8_Analysis}.

%
%

%
%

\begin{figure}[h!]
\centering
  \includegraphics[width=0.99\columnwidth]{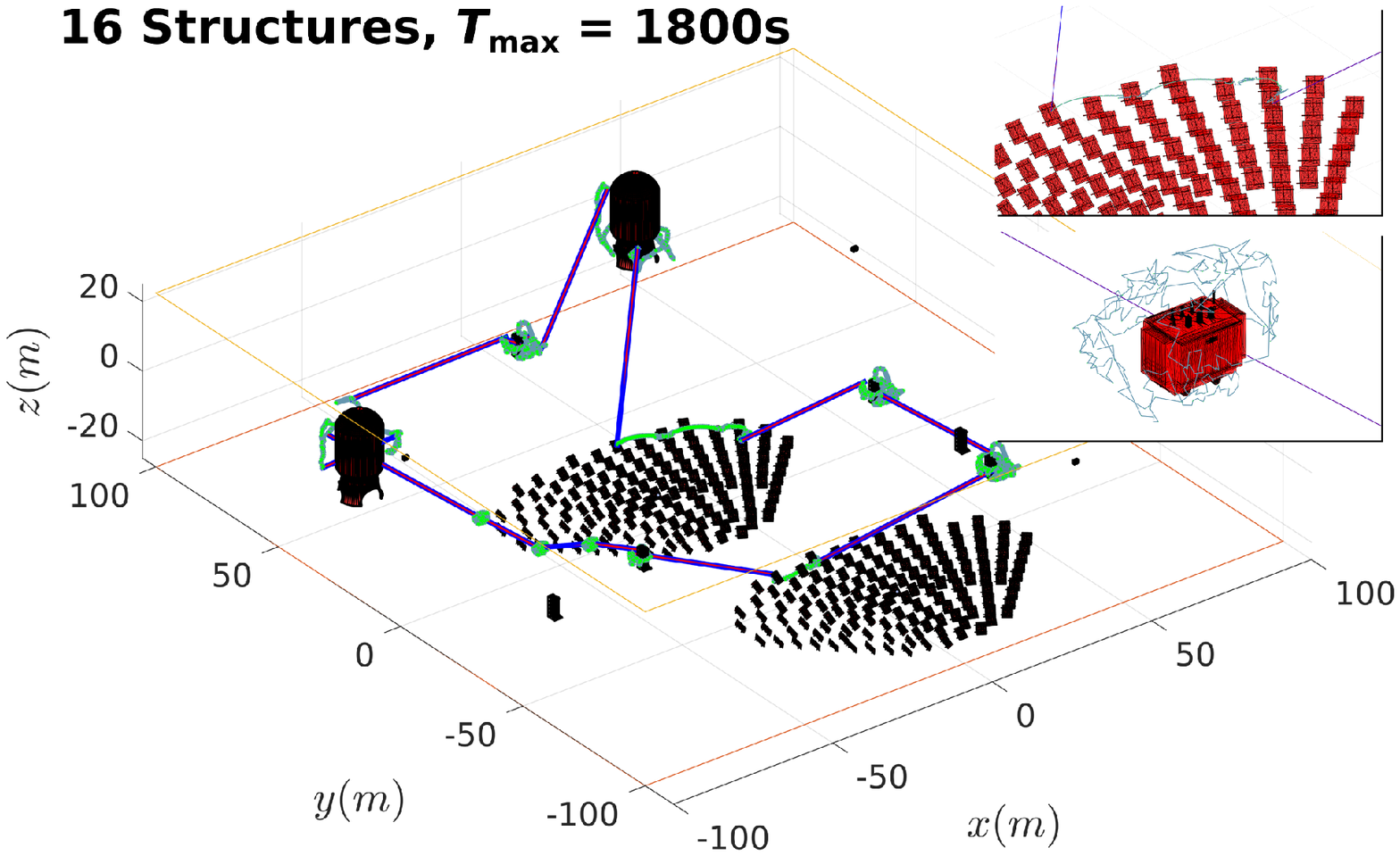}
\caption{Distributed infrastructure inspection result using the proposed algorithm for the case of a large area with $16$ structures and a robot endurance of $T_{\max} = 1800\textrm{s}$. The distribution of the structures is based on a randomized spatial generator.}
\label{fig:DISIP_16}
\end{figure}

\begin{figure}[h!]
\centering
  \includegraphics[width=0.99\columnwidth]{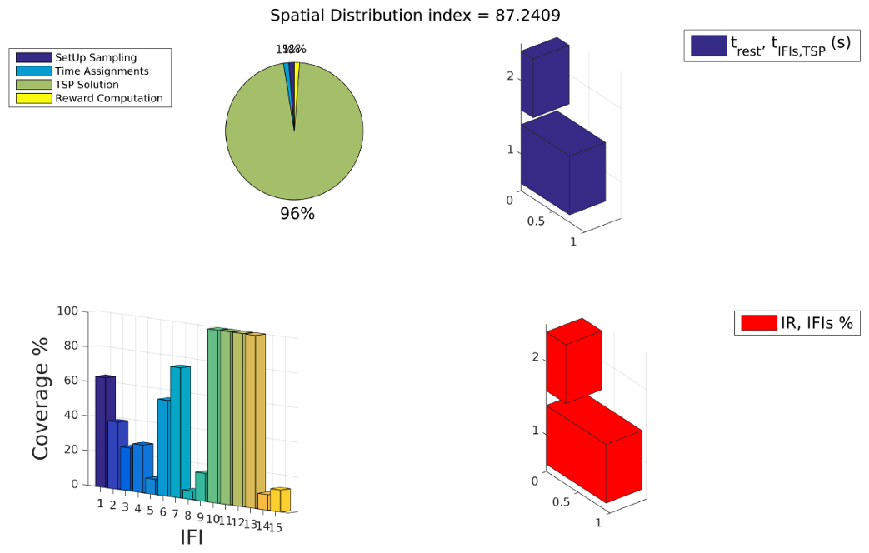}
\caption{Computational analysis of the a distributed infrastructure inspection result using the proposed algorithm for the case of a large area with $16$ structures and a robot endurance of $T_{\max} = 1800\textrm{s}$. The term spatial distribution index denotes the average distance from IFI to IFI. The upper right plot denotes the time for the IFI sampling and TSP computations (lower bar) and all the rest (upper bar). The bottom right plot denotes the percentage of IFIs visited (lower bar) and the percentage of awards collected (upper bar). The upper left pie denotes the percentage of the basic algorithmical step and as shown the TSP solution consumes the majority of the time. Finally, the bottom left plot denotes the percentage of coverage for each of the sampled IFIs. }
\label{fig:DISIP_16_Analysis}
\end{figure}

The second result refers to the same area size and robot endurance but this time populated with $16$ randomly distributed structures. This essentially corresponds to a much denser and more challenging scenario. The derived result, after $k=30$ iterations of the algorithm, is shown in Figure~\ref{fig:DISIP_16}. The relevant statistical analysis is shown in Figure~\ref{fig:DISIP_16_Analysis}.

%
%

%
%

\begin{figure}[h!]
\centering
  \includegraphics[width=0.99\columnwidth]{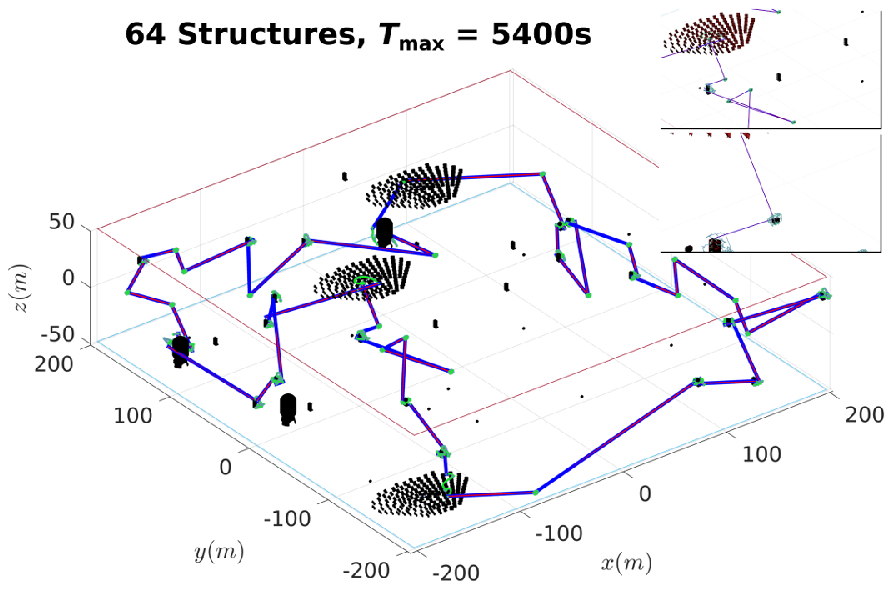}
\caption{Distributed infrastructure inspection result using the proposed algorithm for the case of a large area with $64$ structures and a robot endurance of $T_{\max} = 5400\textrm{s}$. The distribution of the structures is based on a randomized spatial generator.}
\label{fig:DISIP_64}
\end{figure}

\begin{figure}[h!]
\centering
  \includegraphics[width=0.99\columnwidth]{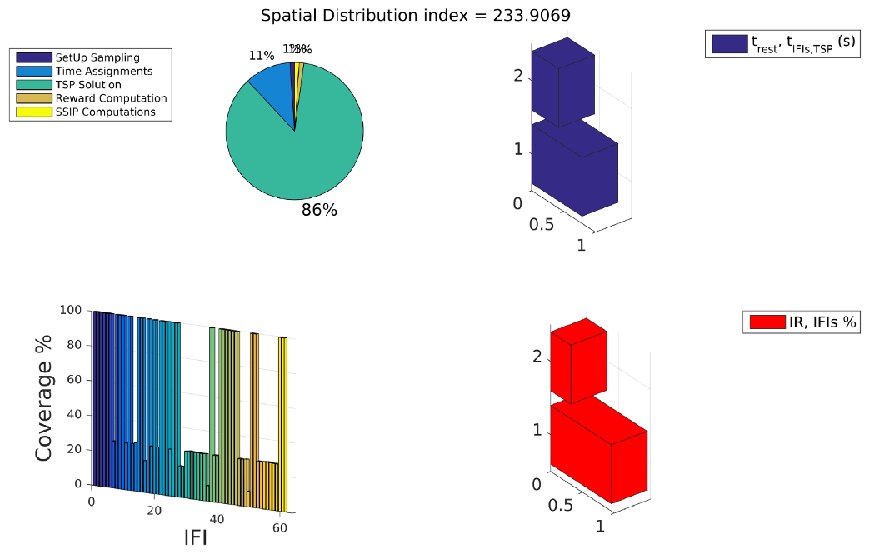}
\caption{Computational analysis of the a distributed infrastructure inspection result using the proposed algorithm for the case of a large area with $64$ structures and a robot endurance of $T_{\max} = 5400\textrm{s}$. The term spatial distribution index denotes the average distance from IFI to IFI. The upper right plot denotes the time for the IFI sampling and TSP computations (lower bar) and all the rest (upper bar). The bottom right plot denotes the percentage of IFIs visited (lower bar) and the percentage of awards collected (upper bar). The upper left pie denotes the percentage of the basic algorithmical step and as shown the TSP solution consumes the majority of the time. Finally, the bottom left plot denotes the percentage of coverage for each of the sampled IFIs.}
\label{fig:DISIP_64_Analysis}
\end{figure}

The third result to be analytically presented refers to the case of an aerial robot with $T_{\max} = 5400\textrm{m}$ operating in a large area populated with $64$ infrastructure facilities. The derived inspection result, after $k=20$ iterations, is shown in Figure~\ref{fig:DISIP_64}. Statistical analysis of the result is presented in Figure~\ref{fig:DISIP_64_Analysis}. In all cases, it is shown that DISIP tends to visit a large percentage of the IFIs, while the computational load per iteration is dominated by the TSP solution - an indication that the additional complexity of the problem is handled efficiently. 

%
%

%
%

\begin{figure}[h!]
\centering
  \includegraphics[width=0.99\columnwidth]{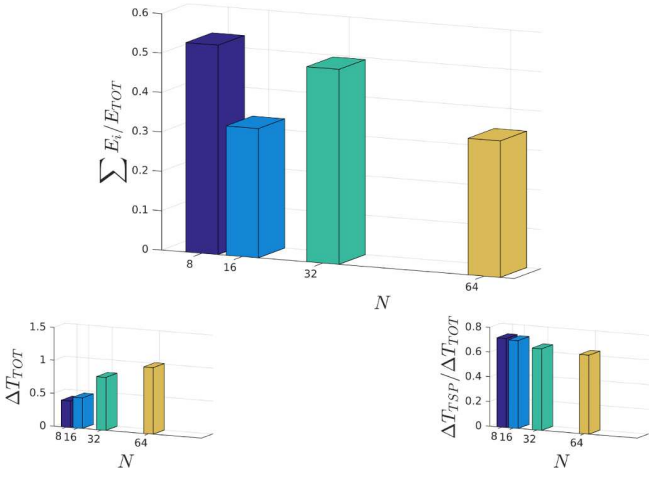}
\caption{Averaged statistics result on the percentage of the overall coverage (top plot), the total computation time per iteration (bottom left) and the ratio of computational load due to the TSP computations (bottom right). Interestingly, it is shown that for the same robot endurance, doubling the amount of IFIs leads to a similar drop of percentage coverage for the two cases of endurance considered. Furthermore, the increase in computational time is almost linear while the TSP computations hold a major and only slowly decreasing percentage of the computations as the amount of IFIs increase. }
\label{fig:DISIP_statistics}
\end{figure}

To provide broader statistical insight on the computational capabilities of the algorithm, such simulations were conducted for $4$ sets, each with $10$ scenarios of $8, 16, 32, 64$ of these structures in smaller and larger areas. The extracted averaged statistical results are presented and discussed in Figure~\ref{fig:DISIP_statistics}, while the complete set of results is available online~\cite{DISIPdatasest}. Note that in this plot the depicted times are per iteration of the DISIP algorithm.

%
%

\subsection{Experimental Studies}\label{subsec:exp}

\begin{figure}[h!]
\centering
  \includegraphics[width=0.95\columnwidth]{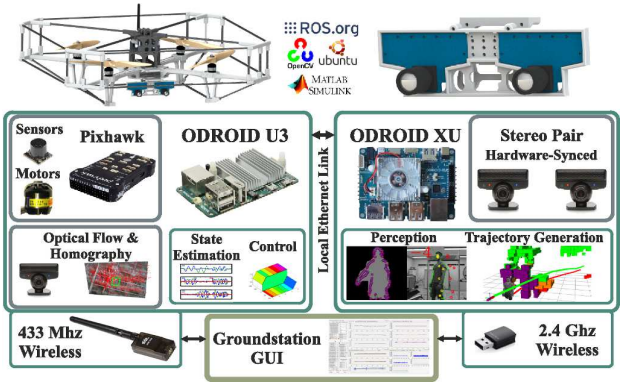}
\caption{Illustration of the autonomous quadrotor UAV employed in experiments and its main hardware components. .}
\label{fig:quad_framework}
\end{figure}

The proposed DISIP algorithm was additionally evaluated experimentally for the case of a simplified, downsized set--up consisting of four structures, namely a power transformer mock--up and assemblies of boxes with different geometries. The distributed inspection mission is conducted using a small aerial robot capable of flying autonomously in GPS--denied environments by relying on an onboard perception module. More specifically, this custom--designed robot is equipped with a Pixhawk ARM M4--based autopilot which is interfaced by an ODROID--U3 Quad--Core embedded computer runing a lightweight linux distribution and acting as a high-level unit.  Apart from interfacing the Pixhawk autopilot, it receives the feeds of a ground--pointed PS3 Eye camera operating at 125Hz. Based on custom--developed Optical Flow and Homography estimation algorithms running at $100$Hz. Via proper data fusion with the IMU updates, the UAV is capable of achieving $6$--Degrees of Freedom pose estimation autonomously. Finally, at the position control level a Model Predictive Control strategy is employed. With this lower--layer of autonomy deployed, this aerial robot further employs a stereo vision pair consisting of two hardware-synchronized PS3 Eye cameras. Using an additional ODROID--XU3 Octa--Core single board computer, the tasks of 3D environment perception, and simultaneous localization and mapping for autonomous navigation, are achieved. Figure~\ref{fig:quad_framework} depicts all the hardware components of the employed aerial robot. 

%
%

\begin{figure}[h!]
\centering
  \includegraphics[width=0.99\columnwidth]{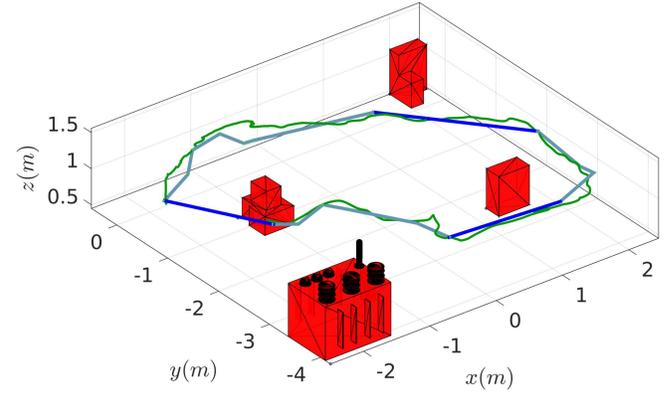}
\caption{Experimental downsized distributed infrastructure inspection scenario with three structures (a power transformer mockup and three sets of boxes) conducted using an autonomous quadrotor navigating in a GPS--denied environment.}
\label{fig:exp_disip}
\end{figure}

\begin{figure}[h!]
\centering
  \includegraphics[width=0.99\columnwidth]{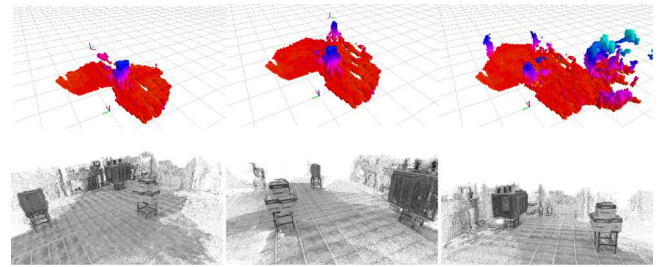}
\caption{Instances of the progressively--built octomap representation of the environmnet which is executed online and in real--time to facilitate autonomous collision--free navigation and views of the offline reconstructed dense point cloud of the inspection scenario and its structures.}
\label{fig:reconstruct}
\end{figure}

Employing this aerial robot, a downsized distributed infrastructure inspection scenario is conducted. Assuming a limited endurance of $T_{\max} = 300\textrm{s}$ and deploying the aforementioned structures on a limited space, DISIP was requested to find a distributed inspection solution that leads to high inspection reward while respecting the endurance of the robot. Note that for this run, the FoV of the PS3 Eye--based stereo sensor ($[40,40]\textrm{deg}$) is respected while a maximum inspection and traveling velocity $v_T = v_T^I = 0.1\textrm{m/s}$ is set in combination with a maximum yaw rate $\dot{\psi}_{\max} = 5\textrm{deg/s}$. Finally, it is highlighted that an open--route (without the final segment to go back to the starting position) was considered here since this: a) makes a better use of the limited set endurance and b) makes sense in this very small space where ``returning--to--home'' is meaningless.  Figure~\ref{fig:exp_disip} depicts the considered scenario alongside with the experimentally recorded trajectory of the robot. In addition, Figure~\ref{fig:reconstruct} shows instances of the online reconstructed octomap voxed--based representation of the environment as well as different views of the dense point cloud derived using postprocessing methods~\cite{Pix4Dsite}. As shown, the real--time computed octomap, which relies on the visual--inertial pipeline of the robot, provides a representation of the environment that facilitates autonomous and collision--free navigation, while the offline reconstructred dense point clouds indicates the sufficient quality of the inspection result for the subsets of the full--coverage path that the DISIP algorithm employed. It is acknowledged that this experimental scenario is not considered to reflect the full-scale of the overall capabilities of the DISIP algorithm due to its small scale, the size of the environment and the endurance of the robot. However, it indicates the applicability of such strategies in terms of inspection quality and handling the vehicle and sensor limitations and models. Furthermore, it indicates the potential provided by autonomously navigation aerial robots when these are combined with intelligent path planning algorithms. 

%
%

%
%

\subsection{Remarks}\label{subsec:remakrs}

The algorithm shows very good scalability properties for different distribution characteristics of the infrastructure. The conducted tests showed that the tendency is to visit a large amount of the infrastructure facilities as long as time is available, even at the cost of covering small percentages of some of them. Detailed analysis of this phenomenon showed that this is because the algorithm can find even short segments of the SSIP paths that however correspond to relatively large coverage percentages. This is not just a feature of the employed SSIP results regarding the viewpoint selection it makes, but a more broad truth as different geometrical complexities of the structure have an impact on the amount of viewpoints and path length required to achieve a level of coverage. Additionally, it is highlighted that the TSP problem -which is one of the conducted steps per iteration-- corresponds to the main computational load. Finally, the experimental study revealed the applicability of the DISIP algorithm.

\section{CONCLUSIONS}\label{sec:concl}

A practically--oriented algorithm that addresses the problem of distributed infrastructure structural inspection path planning subject to time constraints was proposed within this paper. To derive its solution, the proposed approach relies on an iterative, $3$--step optimization paradigm via which it computes first feasible solutions very fast while given that more time is available the solution quality increases. As shown, the algorithm respects the imposed time constraints as well as the vehicle dynamics and sensor limitations. Via an extensive set of simulation scenarios, the path quality, low computational load and scalability regarding the size of the problem and the density of the infrastructure are demonstrated and statistically evaluated. Furthermore, an experimental study using an autonomous quadrotor illustrates the applicability of such paths for inspection operations. A public dataset accompanies this submission, while upon possible acceptance of the paper, the code will be released to the public for further use and development from the community.


\begin{thebibliography}{10}
\providecommand{\url}[1]{#1}
\csname url@samestyle\endcsname
\providecommand{\newblock}{\relax}
\providecommand{\bibinfo}[2]{#2}
\providecommand{\BIBentrySTDinterwordspacing}{\spaceskip=0pt\relax}
\providecommand{\BIBentryALTinterwordstretchfactor}{4}
\providecommand{\BIBentryALTinterwordspacing}{\spaceskip=\fontdimen2\font plus
\BIBentryALTinterwordstretchfactor\fontdimen3\font minus
  \fontdimen4\font\relax}
\providecommand{\BIBforeignlanguage}[2]{{%
\expandafter\ifx\csname l@#1\endcsname\relax
\typeout{** WARNING: IEEEtran.bst: No hyphenation pattern has been}%
\typeout{** loaded for the language `#1'. Using the pattern for}%
\typeout{** the default language instead.}%
\else
\language=\csname l@#1\endcsname
\fi
#2}}
\providecommand{\BIBdecl}{\relax}
\BIBdecl

\bibitem{Metni20073}
N.~Metni and T.~Hamel, ``A uav for bridge inspection: Visual servoing control
  law with orientation limits,'' \emph{Automation in Construction}, vol.~17,
  no.~1, pp. 3 -- 10, 2007.

\bibitem{englot2012sampling}
{B.~Englot, and F. S. Hover}, ``Sampling-based sweep planning to exploit local
  planarity in the inspection of complex 3d structures,'' in \emph{Intelligent
  Robots and Systems (IROS), 2012 IEEE/RSJ International Conference on}.\hskip
  1em plus 0.5em minus 0.4em\relax IEEE, 2012, pp. 4456--4463.

\bibitem{ADBS_AURO_2015}
\BIBentryALTinterwordspacing
K.~Alexis, G.~Darivianakis, M.~Burri, and R.~Siegwart,
  ``\BIBforeignlanguage{English}{Aerial robotic contact-based inspection:
  planning and control},'' \emph{\BIBforeignlanguage{English}{Autonomous
  Robots}}, pp. 1--25, 2015. [Online]. Available:
  \url{http://dx.doi.org/10.1007/s10514-015-9485-5}
\BIBentrySTDinterwordspacing

\bibitem{InfrInspHeli}
T.~Merz and F.~Kendoul, ``Beyond visual range obstacle avoidance and
  infrastructure inspection by an autonomous helicopter,'' in \emph{Intelligent
  Robots and Systems (IROS), 2011 IEEE/RSJ International Conference on}, 2011,
  pp. 4953--4960.

\bibitem{LocLinearStruInsp}
S.~Rathinam, Z.~Kim, and R.~Sengupta, ``Vision-based monitoring of locally
  linear structures using an unmanned aerial vehicle1,'' \emph{Journal of
  Infrastructure Systems}, vol.~14, no.~1, pp. 52--63, 2008.

\bibitem{DABS_ICRA_14}
G.~Darivianakis, K.~Alexis, M.~Burri, and R.~Siegwart, ``Hybrid predictive
  control for aerial robotic physical interaction towards inspection
  operations,'' in \emph{Robotics and Automation (ICRA), 2014 IEEE
  International Conference on}, May 2014, pp. 53--58.

\bibitem{BABOOMS_ICRA_15}
\BIBentryALTinterwordspacing
{A. Bircher, K. Alexis, M. Burri, P. Oettershagen, S. Omari, T. Mantel and R.
  Siegwart}, ``Structural inspection path planning via iterative viewpoint
  resampling with application to aerial robotics,'' in \emph{IEEE International
  Conference on Robotics and Automation (ICRA)}, May 2015, pp. 6423--6430.
  [Online]. Available:
  \url{https://github.com/ethz-asl/StructuralInspectionPlanner}
\BIBentrySTDinterwordspacing

\bibitem{NBVP_ICRA_16}
\BIBentryALTinterwordspacing
{A. Bircher, M. Kamel, K. Alexis, H. Oleynikova and R. Siegwart}, ``Receding
  horizon "next-best-view" planner for 3d exploration,'' in \emph{IEEE
  International Conference on Robotics and Automation (ICRA)}, May 2016.
  [Online]. Available: \url{https://github.com/ethz-asl/nbvplanner}
\BIBentrySTDinterwordspacing

\bibitem{Oettershagen_FSR2015}
{P. Oettershagen, T. Stastny, T. Mantel, A. Melzer, K. Rudin, G. Agamennoni, K.
  Alexis, and R. Siegwart}, ``Long-endurance sensing and mapping using a
  hand-launchable solar-powered uav,'' June 2015.

\bibitem{ZPAT_ISVC_2015}
L.~Zikou, C.~Papachristos, K.~Alexis, and A.~Tzes,
  ``\BIBforeignlanguage{English}{Inspection operations using an aerial robot
  powered-over-tether by a ground vehicle},'' in
  \emph{\BIBforeignlanguage{English}{Advances in Visual Computing}}, ser.
  Lecture Notes in Computer Science.\hskip 1em plus 0.5em minus 0.4em\relax
  Springer International Publishing, 2015, vol. 9474, pp. 455--465.

\bibitem{SIP_AURO_2015}
{A. Bircher, M. Kamel, K. Alexis, M. Burri, P. Oettershagen, S. Omari, T.
  Mantel and R. Siegwart}, ``\BIBforeignlanguage{English}{Three-dimensional
  coverage path planning via viewpoint resampling and tour optimization for
  aerial robots},'' \emph{\BIBforeignlanguage{English}{Autonomous Robots}}, pp.
  1--25, 2015.

\bibitem{APST_MSC_2015}
{K. Alexis, C. Papachristos, R. Siegwart, and A. Tzes}, ``Uniform coverage
  structural inspection path-planning for micro aerial vehicles,'' September
  2015.

\bibitem{bircher_robotica}
{A. Bircher, K. Alexis, U. Schwesinger, S. Omari, M. Burri, and R. Siegwart},
  ``An incremental sampling-based approach to inspection planning: The
  rapidly-exploring random tree of trees,'' 2015.

\bibitem{papachristos2016distributed}
C.~Papachristos, K.~Alexis, L.~R.~G. Carrillo, and A.~Tzes, ``Distributed
  infrastructure inspection path planning for aerial robotics subject to time
  constraints,'' in \emph{2016 International Conference on Unmanned Aircraft
  Systems (ICUAS)}.\hskip 1em plus 0.5em minus 0.4em\relax IEEE, 2016, pp.
  406--412.

\bibitem{vansteenwegen2011orienteering}
P.~Vansteenwegen, W.~Souffriau, and D.~Van~Oudheusden, ``The orienteering
  problem: A survey,'' \emph{European Journal of Operational Research}, vol.
  209, no.~1, pp. 1--10, 2011.

\bibitem{tsiligirides1984heuristic}
T.~Tsiligirides, ``Heuristic methods applied to orienteering,'' \emph{Journal
  of the Operational Research Society}, pp. 797--809, 1984.

\bibitem{blum2007approximation}
A.~Blum, S.~Chawla, D.~R. Karger, T.~Lane, A.~Meyerson, and M.~Minkoff,
  ``Approximation algorithms for orienteering and discounted-reward tsp,''
  \emph{SIAM Journal on Computing}, vol.~37, no.~2, pp. 653--670, 2007.

\bibitem{stanley2001statistical}
H.~E. Stanley and S.~V. Buldyrev, ``Statistical physics: The salesman and the
  tourist,'' \emph{Nature}, vol. 413, no. 6854, pp. 373--374, 2001.

\bibitem{gavalas2015heuristics}
D.~Gavalas, C.~Konstantopoulos, K.~Mastakas, G.~Pantziou, and N.~Vathis,
  ``Heuristics for the time dependent team orienteering problem: Application to
  tourist route planning,'' \emph{Computers \& Operations Research}, vol.~62,
  pp. 36--50, 2015.

\bibitem{yu2014optimal}
J.~Yu, J.~Aslam, S.~Karaman, and D.~Rus, ``Optimal tourist problem and anytime
  planning of trip itineraries,'' \emph{arXiv preprint arXiv:1409.8536}, 2014.

\bibitem{con_dom_sets}
\BIBentryALTinterwordspacing
S.~Guha and S.~Khuller, ``\BIBforeignlanguage{English}{Approximation algorithms
  for connected dominating sets},''
  \emph{\BIBforeignlanguage{English}{Algorithmica}}, vol.~20, no.~4, pp.
  374--387, 1998. [Online]. Available:
  \url{http://dx.doi.org/10.1007/PL00009201}
\BIBentrySTDinterwordspacing

\bibitem{DISIPdatasest}
\BIBentryALTinterwordspacing
{K. Alexis, C. Papachristos}, ``{Distributed Infrastructure Structural
  Inspection Planner Dataset}.'' [Online]. Available:
  \url{https://disip.wikispaces.com/}
\BIBentrySTDinterwordspacing

\bibitem{lin1973effective}
S.~Lin and B.~W. Kernighan, ``An effective heuristic algorithm for the
  traveling-salesman problem,'' \emph{Operations research}, vol.~21, no.~2, pp.
  498--516, 1973.

\bibitem{helsgaun2000effective}
K.~Helsgaun, ``An effective implementation of the lin--kernighan traveling
  salesman heuristic,'' \emph{European Journal of Operational Research}, vol.
  126, no.~1, pp. 106--130, 2000.

\bibitem{LKHcode}
\BIBentryALTinterwordspacing
{Keld Helsgaun}, ``{LKH Version 2.07: an effective implementation of the
  Lin-Kernighan heuristic for solving the traveling salesman problem }.''
  [Online]. Available: \url{http://www.akira.ruc.dk/~keld/research/LKH/}
\BIBentrySTDinterwordspacing

\bibitem{PTT_IROS_15}
C.~Papachristos, D.~Tzoumanikas, and A.~Tzes, ``Aerial robotic tracking of a
  generalized mobile target employing visual and spatio–-temporal dynamic
  subject perception,'' in \emph{Intelligent Robots and Systems, 2015. IROS
  2015. IEEE/RSJ International Conference on}, Sept--Oct 2015.

\bibitem{Pix4Dsite}
{Pix4D}, ``{http://pix4d.com/}.''

\end{thebibliography}


\end{document}